\documentclass{bmvc2k}

 \usepackage{algorithm}
\usepackage{amsmath}
\usepackage{algpseudocode}
 \usepackage{graphicx}
\usepackage{amsfonts}

\title{Real‑Time Feedback and Benchmark Dataset for Isometric Pose Evaluation}

\addauthor{Abhishek Jaiswal}{https://www.cse.iitk.ac.in/users/abhijais/}{1}
\addauthor{ Armeet Singh Luthra}{}{1}
\addauthor{ Purav Jangir}{}{1}
\addauthor{ Bhavya Garg}{}{1}
\addauthor{Nisheeth Srivastava}{}{1}

\addinstitution{
 Indian Institute of Technology Kanpur\\
 Kanpur, India
}

\runninghead{Authors:}{Real-Time Feedback for Isometric Pose Evaluation}

\begin{document}

\maketitle

\begin{abstract}
Isometric exercises appeal to individuals seeking convenience, privacy, and minimal dependence on equipments. However, such fitness training is often overdependent on unreliable digital media content instead of expert supervision, introducing serious risks, including incorrect posture, injury, and disengagement due to lack of corrective feedback. To address these challenges, we present a real-time feedback system for assessing isometric poses. Our contributions include the release of the largest multiclass isometric exercise video dataset to date, comprising over 3,600 clips across six poses with correct and incorrect variations. To support robust evaluation, we benchmark state-of-the-art models—including graph-based networks—on this dataset and introduce a novel three-part metric that captures classification accuracy, mistake localization, and model confidence. 
Our results enhance the feasibility of intelligent and personalized exercise training systems for home workouts. This expert-level diagnosis, delivered directly to the users, also expands the potential applications of these systems to rehabilitation, physiotherapy, and various other fitness disciplines that involve physical motion.
\end{abstract}


\section{Introduction}

Sedentary behaviour is a primary cause of chronic conditions. According to the World Health Organization, one in three individuals is at risk due to lack of physical activity~\cite{who2024physical_activity}. Shifting work cultures with passive work habits has caused daily occupational energy expenditure to decrease by more than 100 calories over the last five decades~\cite{church2011trends}. At the same time, interest in gyms has reduced due to crowded off-working hours, lack of personal assistance, and decline in mobility with age ~\cite{larson2018you,lachman2018adults}. As a convenient alternative, exercisers are looking forward to home workouts~\cite {nwijournal2024virtual_exercise_trends}.

For home exercising, isometric exercises provide ample health benefits without requiring specialized equipments~\cite{wiles2017home,wiles2005novel,baffour2023evidence}. They demand less space constraints owing to the static hold pose, improve general fitness, and are particularly effective when recovering from an injury. While there are plenty of works on motion-based isotonic exercise feedback ~\cite{chen2020pose,wang2019ai,gharasuie2021performance,liu2020posture,10.1145/3604915.3608816}, isometric exercises have not yet received their fair share of research attention.

Isometric workouts in the congenial home settings is mainly guided by online training content from social media platforms.
For example, in practice of traditional yoga poses, eight in ten practitioners depend on a virtual service for their workouts~\cite{doyou2024yoga_survey}. Thus, creating a dependency on online videos for guidance in the absence of a physical trainer~\cite{hinz2021practice,smolenska2021health}.  However, the quality of such content is often unreliable ~\cite{baykara2024recommended,rodriguez2022review}.

Fitness training depending on online content faces two significant drawbacks: First, the exercisers do not get any review of their performance, and Second, there is no real-time exercise engagement, increasing the chances of dropouts. Absence of exercise feedback can cause serious consequences ranging from unsatisfactory outcomes to risk of injury. Additionally, there is no extrinsic motivator like a gym trainer in home exercise, and this lack of feedback can cause enthusiasm deficit diminishing commitment. These shortcomings are inherent to any unsupervised at-home exercise regimes. Therefore, addressing these challenges is crucial for enhancing the quality and sustainability of home-based fitness programs.

Automated feedback for isometric exercises comes with its own unique challenges. While mainstream dynamic exercises are supported by numerous mobile applications offering automated feedback, the same level of refinement is lacking for isometric exercises. One key reason for this gap is the lack of datasets available to the research community. Existing datasets typically include data for the correct pose only, neglecting the common mistakes that are necessary for accurate feedback. Without data on common errors, the go-to feedback mechanism rely on angular thresholds derived from domain knowledge or reference poses \citep{chaudhari2021yog, anand2022yoga}. However, the use of pre-defined thresholds may introduce scaling challenges due to variations in body types and clothing, especially under noisy pose estimation conditions~\citep{armstrong2025validation}. 

Another significant challenge in this domain is the inherent nature of isometric exercises, which involve a substantial hold phase. This phase is characterized by an indefinite pause,
making motion modeling and temporal classification more complex compared to their periodic isotonic counterparts. Additionally, in general, isolating the root cause of mistakes is inherently difficult. Since body joints do not move independently and most movements involve complex, multi-joint interactions, errors can be subtle and multifaceted, complicating the process of automated feedback and correction.

To make progress on the above challenges, we created a multiclass video dataset for isometric exercises, the Isometric-Multiclass Dataset (IMCD), with more than 3600 video clips, encompassing both correct and incorrect isometric poses. These poses were performed by approximately ten different subjects, ensuring a degree of diversity in body types and performance styles. Unlike existing datasets that primarily focus on correct pose execution, the IMCD includes a variety of common mistakes, providing a richer resource for training and evaluating automated feedback systems (Table \ref{tab:isometric-datasets-multiclass}). To benchmark the dataset, we learn a threshold-based classifier and examine other state-of-the-art exercise action recognition models that leverage graph networks. 

Beyond model benchmarking, we also propose a three-part metric to analyze model's confidence in its prediction. This setup is useful for cases when multiple mistakes can co-occur or are interlinked. By focusing on both the identification of the correct pose and the model's confidence level in predicting the incorrect classes, this approach provides a more robust evaluation of model confidence to better analyze the root cause of mistakes. Moreover, with this metric, we can further ensure that the mistake class is not misclassified as correct, which is a necessary safeguard to ensure safety. Such an evaluation becomes critical in scenarios where exercise video datasets are of small size, increasing the risk of deep learning models overfitting to noise.

To summarize, this paper presents the following contributions:
\begin{itemize} 
    \item We release the largest multiclass isometric video dataset of six poses. This dataset replicates situations in natural settings appropriate for home use (Section \ref{dataset}).
    \item We benchmark our dataset with state-of-the-art algorithms from related domains (Section \ref{experiements}).
    \item We present a three-part classification metric to better analyze model performances, taking into account the confidence intervals of model prediction (Section \ref{3part}).

\end{itemize}

\begin{table}[htbp]
\centering
\begin{tabular}{|l|c|l|c|}
\hline
\textbf{Dataset} & \textbf{Type} & \textbf{Source Available} & \textbf{Multiclass} \\
\hline
Yoga-82         & Image & Images      & No \\
YAR             & Video & Video       & No \\
3D-Yoga         & Video & --          & No \\
YogaTube        & Video & --          & No \\
3DYoga90        & Video & Youtube Video   & No \\
IMCD (ours) & Video & Pose CSV    & Yes \\
\hline
\end{tabular}
\vspace{1em}
\caption{ Existing datasets on isometric poses. IMCD: Isometric-Multiclass Dataset}
\label{tab:isometric-datasets-multiclass}

\end{table}

\section{Isometric exercise - Identification, Classification and Feedback}

\paragraph{Pose Recognition:
}Many of the past studies on isometric poses have focused on exercise pose recognition to categorize various types of exercises performed using static images or recorded video datasets
\citep{nagalakshmi2021classification,verma2020yoga,kothari2020yoga,yadav2019real,chen2014yoga,jain2021three,dittakavi2022pose,jain2021three}. 
Pose recognition, even though an important beginning step, does not sufficiently help users lacking feedback while exercising independently. For an exerciser, it is more important to understand when they are making a mistake while performing the yoga pose. 

\paragraph{Single-class Models: 
}Many researchers have developed techniques for identifying correct and incorrect exercise poses using training data from correct pose thresholds~\cite {chaudhari2021yog,chen2018computer,anand2022yoga}. For instance, using an image dataset, \citet{dittakavi2022pose} developed a coarse-to-fine pose recognition framework and an angle likelihood mechanism to identify joints needing correction.
Similarly, \citet{wu2021computer} used contrastive features representation for yoga pose grading, comparing the learner's pose image against the coach's image. For temporal exercise motion, an image-based solution may not be very appropriate as the exerciser input will always be a sequence of actions rather than a single representative hold stance. 
Moreover, thresholds may be suitable for differentiating beginner, intermediate, and advanced levels of an exercise, but manually calibrating them requires domain knowledge, and still, they may not hold with body size variations.
In principle, heuristics could be explored to separate incorrect hold stances using only correct data. Nevertheless, for appropriate biomechanical feedback, knowledge of common exercise mistakes is useful.

Few groups have put together publicly available YouTube videos to develop video-based datasets and benchmarks~\citep{marchetti20243dyogaseg,kim20233dyoga90}.
Developing pose correction models over such datasets is challenging as they primarily contain only correct pose information, that too from inconsistent camera angles. Additionally, different YouTube instructors may have different definitions for the correct pose, introducing ambiguity in the downstream applications. 

\paragraph{Multi-class Models
}More suited to practical use cases are studies involving both correct and incorrect class data. However, existing datasets are not very comprehensive. For example, using a small dataset of planks and holding squats images, \citet{militaru2020physical} developed a CNN-based multi-class classifier with two incorrect classes. For three exercises, \citet{zhao20223d} recorded data using GoPro cameras with four subjects and develop a DCT-based classification and correction model. To cover a broader ranger of general users, we stick to mobile camera devices and record with ten subjects. Using only correct class data, \citet{garg2023short} exploited autoencoder reconstruction error to provide natural language feedback for four rehab exercises. Extending these studies, our work introduces the first large-scale, multi-class, video-based dataset specifically curated for isometric exercises.

\paragraph{Exercise Feedback:} For home settings, it is paramount to test how receptive the users are to various methods of training feedback. To make feedback human-friendly, ~\citet{fieraru2021aifit} provided intuitive natural language explanations by comparing the exercise against a reference from professional instructors. Similarly, previous Human Computer Interaction (HCI) studies have applied innovative ways to provide corrective feedback. 
For instance, \citet{turmo2020bodylights} developed an abstraction of exercise pose using 3D-printed projecting lights that are compared against a designated spatial landmark to set individually tailored benchmarks. For examining the timing of interventions, \citet{shang2019effects} compared within-task and post-task feedback strategies for a sorting task using a VR system. Similarly, \citet{radhakrishnan2020erica} used an in-ear device to deliver feedback and compare three different intervention strategies  - within an exercise set, at the end of an individual set, and after all sets are over. Their results showed that the users preferred within-set feedback.
We provide intermittent rep feedback during the exercise and a detailed poset-exercise assessment to support the user's exercise.

\section{Dataset curation}
\label{dataset}

\begin{figure}[bthp]
    \centering
    \includegraphics[width=.97\linewidth]{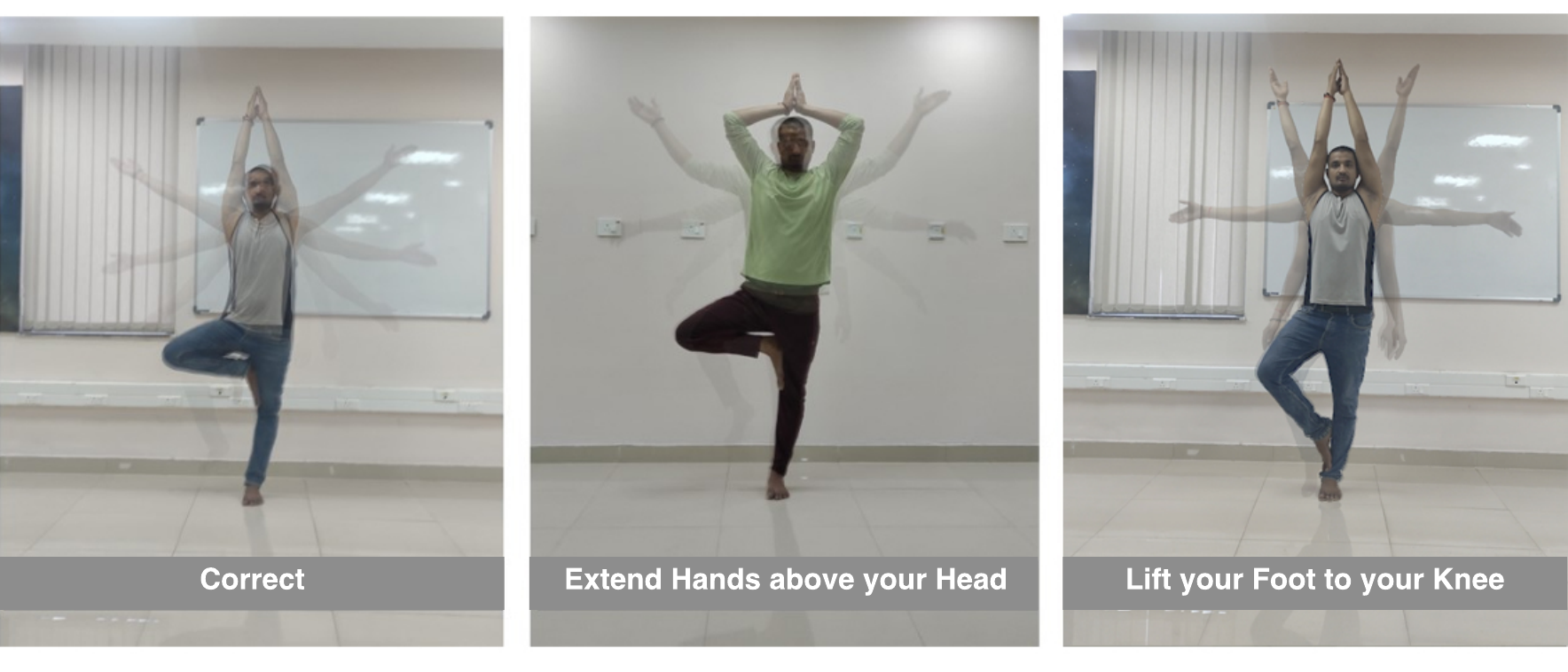}
    \vspace{-1em}
    \caption{A sample from IMCD: Video Motion Trail for Tree Pose for three classes }
    \label{fig:vrik_motion}
\end{figure}

\begin{table}[tbhp]
\centering
\begin{tabular}{|l|c|p{1.2cm}|p{2.5cm}|p{2.0cm}|}
\hline
\textbf{Pose}        & \textbf{Total Reps} & \textbf{Class 1}        & \textbf{Class 2}        & \textbf{Class 3}            \\ \hline
Cobra Pose          & 936                          & Correct                 & Rising on the strength of hands                   & Feet above ground                      \\ \hline
Triangle Pose           & 650                        & Correct            & Right hand not reaching ankle                & Right Knee bending      \\ \hline
Warrior-2 Pose        & 742                          & Correct                 & Hands not parallel to ground      & Right knee bending forward                                   \\ \hline
Plank Pose           & 557                          & Correct                 & Hips too high     & Hips too low          \\ \hline
Tree Pose           & 780                          & Correct                 & Hands not extending above head       & Right foot not reaching left knee                      \\ \hline
Superman Pose         & 674                          & Correct                 & Knees bending             & Hands not at the same level  \\ \hline

\end{tabular}
\vspace{1em}
\caption{Dataset statistics for each isometric pose }
\label{tab:poses_data}

\end{table}

Isometric exercises rely on maintaining static positions (hold phases) and involve dynamic transitions in and out of the static pose. While current datasets focus on hold phase images, they lack the video data that reflects the live scenario of a user entering and exiting the hold phase.  Furthermore, most datasets omit incorrect hold examples, which are critical for training classification models to diagnose errors and provide feedback. While we can check for dissimilarity between the user's pose and a correct reference pose to identify the mistake, such methods fall short of providing actionable biomechanical feedback, lacking knowledge of the common mistakes generally made.

To address these issues, we recorded a comprehensive dataset comprising six exercises, each featuring one correct technique and two commonly made mistakes (Table \ref{tab:poses_data}). Approximately ten male participants of varying body sizes performed each exercise with over 500 reps of correct and incorrect class data. The data was recorded in both indoor and outdoor settings with subjects dressed in standard home attire (for a sample, see Figure \ref{fig:vrik_motion}).

Our objective was to ensure that the dataset closely resembles real-life scenarios, particularly how users capture themselves for the exercise. To achieve this, we utilized various mobile phones placed at different distances while ensuring that the exerciser remained within the frame boundary during the workout. The only restriction enforced was that the mobile camera should be positioned parallel to the exerciser, allowing them to be captured from the front without occlusion. This setting is easily replicable when self-exercising at home. 

Casual phone users were selected to assist with the recording process; they were informed how to position the camera to ensure that the exerciser was fully visible during the exercise. With this straightforward recording setup, we aimed to ensure that the scenario accurately reflects real-life situations in which individuals would set up their own camera. 

The exercisers were given clear instructions on the correct form of an exercise and were corrected to maintain proper technique. For the simulated mistake classes, each exerciser was informed about the specific mistake to demonstrate; however, they had the freedom to simulate these mistakes within the defined parameters of each mistake category. The complete dataset, along with the experiment codes, will be released post-acceptance. A sample of the dataset with experimental code can be found using the following link:
\href{https://drive.google.com/drive/folders/1_2qsXTYfwAFv06e1WGsITrj-cFuznqzH?usp=sharing}{Dataset and Experiments}.

\section{Exercise Classification}
\label{experiements}
The highlight of an isometric exercise is the static hold phase. Hold phase joint angles heavily influence classification, similar to comparing two images. Nonetheless, differences in bodily shape and size, variations in camera positions and angles, and errors in keypoint localization can affect joint angles even for the same correctly performed exercise. Therefore, in our analysis, instead of relying on predetermined heuristic angles from a trainer, we examine the distribution of our dataset to identify the correct range of motion for exercises. We then compare these findings with other state-of-the-art models for exercise classification and compare their performance in live settings for user feedback.

For all our tests, we represent a human pose as a set of keypoints corresponding to body joints. Given a skeleton with $N$ joints, we denote $p_i = (x_i, y_i)$ as the 2D coordinates of the $i$-th joint. This representation is common for all our examined methods.

\paragraph{ Framework for Rep Segregation.}
To detect exercise repetitions (reps) for subsequent analysis, we begin by examining the motion data time series for a specific joint, represented as $\mathbf{y} = [y_1, y_2, \ldots, y_n]$ showing periodicity in motion. We identify local maxima, denoted as $\mathcal{P}_k$, in this signal that correspond to the hold phase, where $P_k$ indicates the peak value at index $k$. The prominence $\Delta P_k$ of a peak measures how much the peak stands out relative to the surrounding peaks around $P_k$~\cite{wikipediaTopographicProminence}. The prominence is defined as:

\begin{equation}
\Delta P_k = y_k - \min(y_i, y_j) \quad y_i \text{ and } y_j \text{ are neighboring points in the valleys around peak } P_k 
\label{eqn:peak_value}
\end{equation}

We only retain those peaks for which the prominence $\Delta P_k$ exceeds a predefined threshold ($\tau = 0.2$). This indicates that the peak represents a significant movement transition.

After identifying the prominent peaks, we determine the exact start and end points of each repetition by analyzing the signal around each peak. For a peak at position $P_k$, we inspect a predefined region surrounding the peak to identify the start and end points as the first local minimum preceding and following $P_k$, respectively.

Once we have identified the repetitions in this manner, we can use these time series segments for exercise classification.

\subsection{Model 1 - Angle-based classifier}
\paragraph{Feature Extraction}
For each rep, angle $\theta_{ijk}$  between joint triplets $(i,j,k)$ is computed as:
\begin{equation}
\theta_{ijk} = \mathrm{atan2}\left(
    \frac{\vec{v}_{ij} \cdot \vec{v}_{jk}}{\|\vec{v}_{ij}\| \|\vec{v}_{jk}\|}
\right), \qquad \theta_{ijk} \in [0^\circ, 360^\circ], \text{post transformation}
\end{equation}
where $\vec{v}_{ij} = (x_j{-}x_i,\, y_j{-}y_i)$ and $\vec{v}_{jk} = (x_k{-}x_j,\, y_k{-}y_j)$ are joint vectors.

For each $\theta_i$ in the time series, a histogram is computed over $n$ bins spanning $[0^\circ, 360^\circ]$:
\begin{equation}
H_{\theta_i}(b_k) = \sum_{t=1}^N \mathbb{I}\big(B_k \leq \theta_i^{(t)} < B_{k+1}\big)
\end{equation}
where $B_k$ are bin edges and $N$ is the number of pose samples. $H_{\theta_i}(b_k)$ represents the count of angles in $k$-th bin.

The feature vector $\mathbf{f}$ consists of the bin centers with the highest counts for each angle:
\begin{equation}
\mathbf{f} = \big[
    \underset{B_k}{\mathrm{arg\,max}}\, H_{\theta_1}(b_k),\;
    \ldots,\;
    \underset{B_k}{\mathrm{arg\,max}}\, H_{\theta_m}(b_k)
\big]
\end{equation}

This feature vector captures the predominant angular patterns, allowing for the classification of pose correctness through learned discriminative features.

Many previous studies have focused on manually defining thresholds for correct body angles based on a trainer's reference. However, such methods may struggle from bodily variations. In contrast, we take a data-driven approach to prototype a correct pose. From this prototypical pose, we can deviate in different degrees to mark correct exercises at various difficulty levels. For standard assessments, we consider the range of angles within one standard deviation above and below the mean as correct. This range can also be adjusted to create categories for beginner, advanced, and professional levels of correct exercise, similar to the work of~\citet{wu2021computer} on pose grading.

Using data from all the exercise classes for multiple exercise-specific angles, we trained various classifiers in a multi-class setting. We found the multi-layer perceptron (MLP) classifiers to be the most robust for our data.

Since this solution does not rely on complicated machine learning architecture, it can provide feedback in near real-time, allowing users to promptly correct their exercise form and avoid potential injuries, while also taking into account the temporal nature of exercise data.

\subsection{Model 2 - Physics-based motion predicting model}

Based on the work of \citet{10.1145/3604915.3608816}, we evaluate a motion forecasting model trained exclusively on correct exercise data to differentiate exercise pose. For any given exercise, the difference in actual motion performed and the model prediction guides exercise classification. Even for a stationary stance, the posture deviations reveal significant differences, especially in the case of incorrectly positioned joints. We leverage these deviations, transformed into the frequency domain, to train a posture classifier.

We focus on physics-predicting models as they provide a natural representation to define per-frame joint position errors. This motion prediction also accounts for variations in individual body types, thereby avoiding a one-size-fits-all solution. For such a pipeline, any model that can accurately predict motion rollout can serve as a motion forecasting model. In this study, we use Interaction Networks~\cite{battaglia2016interaction} as our motion forecasting model and adopt a discrete-time Fourier transform-based method for deviation classification, following the approach described by \citet{10.1145/3604915.3608816}. However, it is worth noting that other variable-length time series classification methods may also be explored. For more details about the model, readers can refer to the original pipeline~\cite{10.1145/3604915.3608816}.

\subsection{Model 3 - Two-Stream Adaptive Graph
Convolutional Networks(2s-AGCN)}
Graph Networks are naturally suited to model skeleton data. Spatiotemporal models connect the body joints in each frame spatially while also linking corresponding joints across different video frames to incorporate temporal information. Building on the SpatioTemporal Graph Convolution Networks(STGCN) \cite{yan2018spatial}, many different variations have been proposed in different action recognition settings \cite{sun2022human,liu2020disentangling,shi2019two,jaiswal2024benchmarking}. In our work, we focus on Two-Stream Adaptive Graph Convolutional Networks (2s-AGCN) \cite{shi2019two} for comparative analysis.

The 2s-AGCN model \cite{shi2019two} combines first-order information (joint coordinates) and second-order information (bone length and direction) over dual graphs. It adaptively learns the graph's topology through back-propagation. The model develops an adaptive joint adjacency matrix in an end-to-end manner to effectively capture spatial relationships specific to the data. Ultimately, scores from the joint stream and the second-order bone vector stream are fused to make the final prediction.

While 2s-AGCN and its variants perform well for action recognition, not all are suitable for real-time exercise feedback. Nevertheless, some adaptations have shown promising results in real-time settings \cite{parsa2020spatio,duan2022pyskl}, although further exploration is needed to understand their applicability for fitness training. In our study, we focused on 2s-AGCN for its simplicity, demonstrating that action recognition models can be effectively employed for isometric pose recognition.

\section{Results}

\begin{table}[h!]
\centering

\begin{tabular}{|l|c|c|c|}
\hline
\textbf{Pose}        & \textbf{Angle-based (norm.)} & \textbf{Motion prediction} & \textbf{2s-AGCN} \\ \hline
Cobra Pose          & \textbf{0.952}                        & 0.824         & 0.914          \\ \hline
Triangle Pose           & \textbf{0.894}                        & 0.563         & 0.856           \\ \hline
Warrior-2 Pose        & \textbf{0.796}                        & 0.669         & 0.772           \\ \hline
Tree Pose           & 0.947                        & 0.837         & \textbf{0.956}           \\ \hline
Plank Pose           & \textbf{0.932}                        & 0.610          & 0.931           \\ \hline
Superman Pose          & 0.802                         & 0.701          & \textbf{0.830}           \\ \hline
\end{tabular}
\vspace{1em}
\caption{Weighted F1 scores based performance comparison of pose classification methods on the Isometric-Multiclass Dataset (IMCD)}
\label{tab:pose_classification}

\end{table}

Table \ref{tab:pose_classification} highlights the efficacy of using an angle-based classification technique over other motion-based methods. These results align with the dataset's focus on hold phase errors, where deviations in joint positions during the hold phase directly correspond to pose inaccuracies.
Notably, 2s-AGCN slightly outperforms Angle-based methods for Tree pose and Superman pose, suggesting that its adaptive graph topologies capture pose subtleties even in predominantly static scenarios. On the other hand, the Motion-prediction model consistently performed poorly on our dataset, underscoring its limitation in static pose error detection. 


\section{Three-Part Metric}
\label{3part}
\begin{table}[h!]
\centering
\begin{tabular}{|l|c|c|c|c|}
\hline
\textbf{Method}    & \textbf{Multiclass F1}   & \textbf{M1 (F1 Binary)}&   \textbf{M2 (F1)} & \textbf{M3(in\%)} \\ \hline
Motion based   &   0.837        & 0.801                & 0.868       & 3.60        \\ \hline
2s-AGCN  &  \textbf{0.956 }   & 0.870               & 0.976       & 12.9      \\ \hline
Angle-based   &  0.947     & \textbf{0.959 }               & 0.972       &\textbf{ 0.21 }       \\ \hline

\end{tabular}
\vspace{0.2em}

\caption{Demonstration comparison using the three-part metric for Tree Pose }
\label{tab:3part_metric}
\end{table}

Exercise mistakes are often interrelated, where errors in one joint often compound others, increasing uncertainty in model predictions. Traditional metrics fail to address this ambiguity, particularly for co-occurring or overlapping mistake types. To resolve this, we propose a three-part metric that explicitly quantifies prediction uncertainty(Algorithm~\ref{alg:3part_metric} ):

\textbf{M1} evaluates the model’s ability to identify correct and incorrect poses, measured via binary F1 score.

\textbf{M2} covers those examples where the model identifies a mistake with a certain degree of confidence (say, prediction probability greater than 50\%).

\textbf{M3} captures low-confidence mistakes, where the model is unsure of the mistake category. 

We can also replace the standard F1 score with $F_\beta$ to prioritize precision to avoid classifying an incorrectly done rep as correct, which might be preferred for exercise settings.

The results from Table~\ref{tab:3part_metric} show that while 2s-AGCN achieves a higher multiclass F1 score, the angle-based method performs better under the three-part metric evaluation. Notably, 2s-AGCN exhibits a significantly higher uncertainty (M3) and lower binary classification scores (M1), indicating that despite its superior multiclass performance, it struggles with prediction confidence and binary task accuracy. In contrast, the angle-based approach achieves the best binary F1 score and demonstrates exceptional reliability with minimal uncertainty. 
Overall, these results highlight the relevance of the proposed three-part metric over the standard multiclass F1 score. In particular, the metric helps identify models that not only classify accurately but also do so with high confidence—a crucial requirement in exercise and rehabilitation settings where ambiguous feedback could lead to user confusion or even injury. Thus, while advanced models like 2s-AGCN may excel in capturing complex error patterns, simpler angle-based approaches may offer superior reliability and safety for real-world deployment.

\begin{algorithm}
\caption{Three-Part Exercise Assessment Metric}
\label{alg:3part_metric}

\begin{algorithmic}[1]
\Require Predictions $P$, Ground truth $Y$, Threshold $\tau=0.5$
\Ensure Metrics $M_1$, $M_2$, $M_3$

\Statex \textbf{Part 1: Correct Class Identification (M1)}
\State Filter examples where $Y_i$ is correct or $P_i$ predicts correct
\State Calculate $M_1 \gets \text{F1-score}(Y_{\text{filtered}}, P_{\text{filtered}})$

\Statex \textbf{Part 2: Incorrect Class Resolution (M2)}
\State Filter: $Y_i$ incorrect $\land$ $\arg\max P_i \neq \mathrm{correct}$ $\land$ $\max P_i^{\mathrm{(incorrect)}} \geq \tau$
\State Calculate $M_2 \gets \text{F1-score}(Y_{\text{filtered}}, P_{\text{filtered}})$

\Statex \textbf{Part 3: Uncertainty Quantification (M3)}
\State Filter: $Y_i$ incorrect $\land$ $\arg\max P_i \neq \mathrm{correct}$ $\land$ $\max P_i^{\mathrm{(incorrect)}} < \tau$
\State $M_3 \gets \frac{\text{Count(filtered examples)}}{\text{Total incorrect examples}}$

\State \Return $M_1$, $M_2$, $M_3$
\end{algorithmic}
\end{algorithm}

\section{Discussion and Limitations}

No study on isometric pose classification could be truly complete without a proper benchmarking dataset, particularly for home exercise settings where models must perform reliably in natural environments and provide real-time predictions. To address these needs, we introduce the first large-scale multi-class isometric pose dataset. Our dataset and the benchmarking approach with a new metric address critical gaps in the existing literature, notably the absence of incorrect pose examples and the necessity for robust evaluation protocols tailored to fitness training. By collecting data from diverse subjects in different environments, we aim to ensure that models trained on this dataset are broadly applicable, thereby supporting the overarching goal of delivering accessible, technology-driven fitness guidance. 

For benchmarking, we focus on an angle-based method incorporating hold phase identification, which facilitates rapid feedback for users. While angle-based methods are straightforward, they have proven effective for classifying isometric poses. Graph-based models show promise in pose classification, but have not yet been adequately explored for exercise feedback.  

 Despite these advances, several limitations remain. Our dataset, while comprehensive, is restricted to six poses and a limited set of common mistakes. The participant pool consists solely of males, which may affect generalizability for other demographics. Furthermore, user studies are needed to examine long-term adherence and reliability of such systems. Furthermore, we need more nuanced feedback mechanisms to meet the needs of the growing society. On the modeling front, our exploration of action recognition models has thus far been limited—a promising direction for future research. In conclusion, expanding mistake categories, diversifying the participant pool, and conducting user studies will be essential for evaluate the broader impact of such systems and should guide the way ahead.

\bibliography{egbib}
\end{document}